# Fully Non-Homogeneous Atmospheric Scattering Modeling with Convolutional Neural Networks for Single Image Dehazing

Cong Wang, Yan Huang, *Member, IEEE*, Yuexian Zou*, *Senior Member, IEEE*
and Yong Xu, *Senior Member, IEEE*

*Abstract*—In recent years, single image dehazing models (SIDM) based on atmospheric scattering model (ASM) have achieved remarkable results. However, it is noted that ASM-based SIDM degrades its performance in dehazing real world hazy images due to the limited modelling ability of ASM where the atmospheric light factor (ALF) and the angular scattering coefficient (ASC) are assumed as constants for one image. Obviously, the hazy images taken in real world cannot always satisfy this assumption. Such generating modelling mismatch between the real-world images and ASM sets up the upper bound of trained ASM-based SIDM for dehazing. Bearing this in mind, in this study, a new fully non-homogeneous atmospheric scattering model (FNH-ASM) is proposed for well modeling the hazy images under complex conditions where ALF and ASC are pixel dependent. However, FNH-ASM brings difficulty in practical application. In FNH-ASM based SIDM, the estimation bias of parameters at different positions lead to different distortion of dehazing result. Hence, in order to reduce the influence of parameter estimation bias on dehazing results, two new cost sensitive loss functions, β-Loss and D-Loss, are innovatively developed for limiting the parameter bias of sensitive positions that have a greater impact on the dehazing result. In the end, based on FNH-ASM, an end-to-end CNN-based dehazing network, FNHD-Net, is developed, which applies β-Loss and D-Loss. Experimental results demonstrate the effectiveness and superiority of our proposed FNHD-Net for dehazing on both synthetic and real-world images. And the performance improvement of our method increases more obviously in dense and heterogeneous haze scenes.

## I. INTRODUCTION

SINGLE image dehazing aims to recover the clean image from a hazy input, which has received significant attention in the vision community over the past few years. According to the atmospheric scattering model (ASM) [1, 2, 3], the hazing process is usually formulated as:

$$I(x,y) = J(x,y)t(x,y) + A\big(1 - t(x,y)\big) \qquad (1)$$

C. Wang, Y. Huang, and YX. Zou are with the ADSPLAB, School of Electronic and Computer Engineering, Peking University, Shenzhen, China (E-mail: congwang22@pku.edu.cn; aihuangy@gmail.com; zouyx@pku.edu.cn).

Y. Xu is with the School of Computer Science and Engineering, South China University of Technology, Guangzhou, China (E-mail: yxu@scut.edu.cn).

YX. Zou and Y. Xu are also with Peng Cheng Laboratory, Shenzhen, China.
*Corresponding Author: YX. Zou.

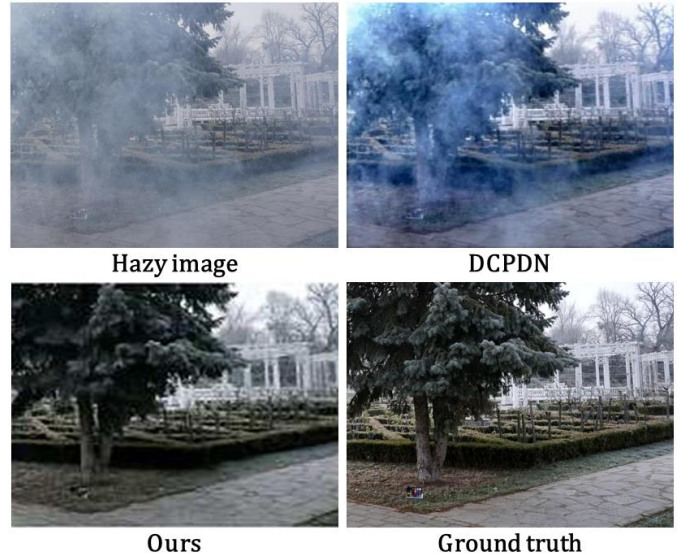

Hazy image     DCPDN

Ours     Ground truth

**Fig. 1.** The dehazing results with different methods. Our method can get clean results without haze residue and artifacts.

$$t(x,y) = e^{-\beta d(x,y)} \qquad (2)$$

where $I(x,y)$ is the hazy image. $J(x,y)$ is the haze-free image. $\beta$ is the angular scattering coefficient (ASC), which indicates haze concentration. The denser haze scene should have a bigger ASC. $A$ is the atmospheric light factor (ALF). $\beta$ and $A$ can be considered as constants among the whole image. $d(x,y)$ is depth. $t(x,y)$ is transmission. Given a hazy image, most dehazing methods try to estimate $t$ and $A$, then calculate clean image by ASM.

However, it has been observed that the ALF is not constant under all the conditions, such as there are both shadow and bright areas in one image. Motived by that, Wang [4] modifies ASM by correcting ALF to be a position dependent variable and proposes a non-homogeneous atmospheric scattering model (NH-ASM). The form of NH-ASM can be expressed as:

$$I(x,y) = J(x,y)t(x,y) + A(x,y)\big(1 - t(x,y)\big) \qquad (3)$$

where $t(x,y) = e^{-\beta d(x,y)}$, which is the same as ASM. NH-ASM has one improvement over ASM, which is modelling ALF as position dependent. Based on NH-ASM, Wang [4]

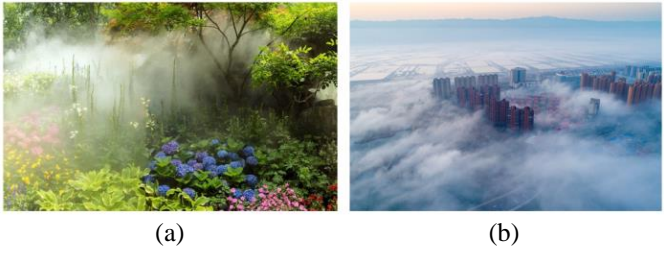

| (a) | (b) |

**Fig. 2.** Two typical hazy scenes with heterogeneous haze degrees: **(a)** Mass haze in the deep forest. **(b)** Advection fog.

builds a CNN [5] module to estimate transmission $t$ and ALF $A$, and calculate clean image $J$ by (4):

$$J(x,y) = \frac{I(x,y) - A(x,y)}{t(x,y)} + A(x,y) \qquad (4)$$

Although NH-ASM has effectively improved ASM, it is still unable to fully describe the complex haze phenomenon in the real world. In NH-ASM, the angular scattering coefficient (ASC) $\beta$ is still a constant, but it has also been observed not a globally constant under any conditions, such as mass haze and advection fog [6]. The former often occurs in deep forests and valleys, and the latter can be seen in the coastal cities of eastern China and western America in spring and winter. The images of such two scenes is shown on Fig. 2. This deviation in modeling causes some biases in dehazing results. To modeling more general dehazing scenes with heterogeneous haze degrees, we modify NH-ASM and correct ASC to be related to pixel position on the image. This model is termed as fully non-homogeneous atmospheric scattering model (FNH-ASM), as all the parameters in the model are position dependent.

Using FNH-ASM based module to dehaze, it needs to estimate 3 parameter maps: ASC, depth and ALF. To reduce the influence of FNH-ASM parameter estimation bias on final dehazing results, in this paper, we explore the correlations between the bias of FNH-ASM parameters and the bias of dehazing outputs and utilize them to restrain dehazing distortion.

We analyze FNH-ASM and get some interesting discoveries (details are elaborated in section III. A):

1) It is a linear relationship between the bias of ALF and the bias of dehazing result, but the effect of the bias of ASC or depth on the bias of final result is nonlinear;

2) The positive and negative of ALF estimation bias has not been seen effects on dehazing result, but it is different for ASC and depth. Compared with the negative bias of ASC and depth, positive bias causes severer biases on dehazing outputs;

3) When the estimation bias is the same, the value of ALF itself has not been seen effects on the bias of dehazing results. However, the value of ASC and depth themselves will affect the deviation of dehazing results. In a heavy hazy image patch, the value of at least one of ASC and depth will be very large. In this case, the bias of ASC and depth have greater impacts on the dehazing result and cause artifacts more easily.

Based on the above discoveries, as the positive and negative bias of estimated ALF do not need to be treated differently, it does not need to set cost sensitive loss function as the optimization objective for training ALF estimator. In other words, the loss functions with uniform penalties or deviation are suitable for ALF estimation, such as MSE loss function [7] or FWB-Loss [4]. However, the loss function for estimating ASC and depth should have the ability to impose heavier penalties on the positive bias and the bias when the value of ASC and depth themselves are larger. Based on such characteristics, it is suitable for training ASC and depth estimators using cost sensitive loss functions. Moreover, ASC and depth are only the intermediate variables in the process of dehazing, and the ultimate optimization goal of the whole dehazing model is minimizing the bias of dehazing results, not just making the bias of ASC or depth minimized. Therefore, it needs to substitute the bias of dehazing result into FNH-ASM to establish the relationship with the bias of estimated ASC and depth, so as to give the designed ASC and depth optimizing loss function the ability to minimize the bias of dehazing result. Based on the above motivations, we design two exponential cost sensitive loss functions, termed as β-Loss and D-Loss, as the optimization goal of ASC and depth estimating module, respectively. β -Loss can simultaneously impose heavier penalties on the positive deviation of ASC, as well as inflict greater punishment on the bias of estimated ASC when the ASC itself is large, and minimize the bias of final dehazing result. Just as β -Loss plays the role in the optimization of ASC estimator, D-Loss can also play the same role in the optimization of depth estimation.

Based on FNH-ASM, we propose a U-Net-based [8, 9], end-to-end single image dehazing network, termed as fully non-homogeneous dehazing network (FNHD-Net). The network can estimate ALF, ASC and depth pixel by pixel. Moreover, for restraining the deviation of dehazing results, FNHD-Net focuses on restricting the positive bias of ASC and depth as well as giving the bias of large ASC and depth values greater penalties. Experiments show that our method is capable of removing haze residue and achieving a better dehazing effect in both homogeneous and heterogeneous haze scenes. The major contributions of our work are:

1) We propose a fully non-homogeneous atmospheric scattering model (FNH-ASM), which models angular scattering coefficient (ASC) as related to pixel position in an image. It can express the imaging law of heterogeneous haze better. Moreover, we find out the cause of artifacts and haze residue on dehazing results by analyzing the correlation between the bias of FNH-ASM parameters and the bias of dehazing outputs.

2) We propose two new cost sensitive exponential loss functions for training ASC and depth estimating module, termed as β-Loss and D-Loss, respectively. They can impose greater penalties on the parameter estimation bias that have great influences on the final outputs to get high-quality dehazing results without haze residue.

3) Based on FNH-ASM, we propose an end-to-end dehazing convolutional neural network, termed as fully non-homogeneous dehazing network (FNHD-Net), which applies β-Loss and D-Loss. The experimental results on both synthetic and real-world hazy databases provide strong support for the effectiveness of our proposed method.

## II. RELATED WORK

In this section, we will discuss the existing single image dehazing algorithms, FWB-Loss and U-Net architecture. The single image dehazing methods can be further divided into two categories: image prior-based methods and deep learning-based methods.

### A. Single image dehazing

Single image dehazing methods have progressed extensively. These methods are roughly classified into two categories: image prior-based methods and deep learning-based methods.

**Image prior-based methods** try to find prior information that can be definitive in improving the visibility of hazy images, such as dark channel prior (DCP) [10], color attenuation prior (CAP) [11] and non-local prior [12]. Although these prior-based methods achieve promising results, the priors depend on the relative assumption and specific target scene, which leads to less robustness in the complex practical scene. For instance, DCP [10] cannot well dehaze the sky regions, since it does not satisfy the prior assumption.

**Deep learning-based methods** try to estimate parameters of ASM or restore haze-free images directly by neural networks. DehazeNet [13] and MSCNN [14] propose multi-scale CNNs [5] for improving the effect of feature extraction and estimating transmission delicately. AOD-Net [15] and IASM-Net [16] take the lead in construct an end-to-end neural network based on the transformed ASM, and realize image dehazing by learning and estimating ASM parameters. After that, various end-to-end methods [4, 17, 18, 19, 20, 21, 22, 23] have been proposed to directly learn hazy-to-clear image translation. Specifically, FWB-Net [4] can effectively restrain color shift of dehazing results via specially designed non-homogeneous atmospheric scattering model (NH-ASM) and loss function. GFN [17], GDN [18], FFA-Net[19] and MSBDN [20] propose novel feature fusion and attention strategies and use them to builds a gated fusion network, a grid network, a feature fusion attention network and a boosted network for single image dehazing, respectively. CPDN [21], EPDN [22] and FD-GAN [23] use generative adversarial network (GAN) [24, 25] to achieve dehazing. Their typical pipeline is using generators to produce haze-free images directly from hazy images, and judge the authenticity of the dehazed image by discriminators.

However, due to the modeling defects of physical models and model estimation bias, both deep learning-based methods and image prior-based methods cannot avoid bias in dehazing when the degrees of haze are different in the whole image. Those biases cause artifacts and haze residue on dehazing results.

### B. FWB-Loss

Front white balance loss function (FWB-Loss) [4] is a loss function for training atmospheric light factor (ALF) estimating module. It can reduce color shift of dehazing results by transforming ALF from RGB to CIELAB color space and minimizing the bias of estimated ALF maps in CIELAB color space. The form of FWB-Loss is:

$$\text{Loss}(A_{est}, A_{gt}) = \sum_{x=1}^{M}\sum_{y=1}^{N}\sum_{k}\left[A_{est}(x,y,k) - A_{gt}(x,y,k)\right]^2 \quad (5)$$

where $A_{est}$ and $A_{gt}$ mean the estimated and ground truth ALF, respectively. $k \in \{L, a, b\}$ means the data channel of CIELAB color space.

### C. U-Net architecture

U-Net [8, 9] is a convolutional neural network architecture that has been applied and shown outstanding performance to many computer vision tasks, such as image segmentation [26, 27, 28], image super-resolution [29], image dehazing [30, 31, 32] and so on.

The classical architecture of U-Net is composed of a contracting path and an expanding path, which correspond to an encoder and a decoder, respectively. Specifically, as an encoder-decoder network, U-Net architecture can be divided into 4 parts: convolution operations, down-sampling operations, up-sampling operations and concatenation operations. The contracting path consists of repeated operations of convolution and down-sampling layers. Symmetrically, the expanding path is made up of convolution and up-sampling layers. To fuse more context information, concatenation operations are adopted between contracting and expanding paths at the same depth level.

Compared with other CNN models, U-Net not only can realize image features with a multi-scale recognition and fusion and obtain high-quality pixel-level output results, but also has a simple and flexible structure that can adapt to new tasks easily. All these advantages reflect the progress and superiority of U-Net architecture.

## III. PROPOSED METHOD

To solve the problem of haze residue in dehazing results, in this section, we first propose a new hazy imaging physical model, fully non-homogeneous atmospheric scattering model (FNH-ASM), to model heterogeneous haze scenes accurately. Then, we obtain a series of strategies to restrain the haze residue of dehazing results by analyzing the relationship between dehazing result bias and the model parameter estimating bias of the dehazing algorithm based on FNH-ASM. Next, based on the strategies above, we propose two cost sensitive loss functions, β-Loss and D-Loss, to limit the haze residue in dehazing results. Finally, we propose a non-homogeneous dehazing network (FNHD-Net), which applies β-Loss and D-Loss in the training phase, and can suppress the haze residue in dehazing results effectively.

### A. Fully non-homogeneous atmospheric scattering model

The classic form of atmospheric scattering model (ASM) is shown in (1) and (2). In ASM, the atmospheric light factor (ALF) $A$ and angular scattering coefficient (ASC) $\beta$ are two global constants. It is suitable for images taken in open fields under sunlight. But when there are shadows or multiple light sources in one scene, the optical property of different pixels in the image are distinct. Therefore, Wang [4] modifies ASM by changing the atmospheric light factor as position dependent and proposes a non-homogeneous atmospheric scattering model (NH-ASM), which has more versatility and robustness than ASM on haze modeling. However, in NH-ASM, ASC is still a constant, which is only suitable for the haze with a uniform concentration. But for the heterogeneous haze, such as mass haze in deep forest or valley and advection fog in the city [6], ASC should be modeled as position dependent. So, we modify NH-ASM and represent ASC as a pixel-related variable. The new model is termed as fully non-homogeneous atmospheric scattering model (FNH-ASM), as all the parameters in the model are position dependent. FNH-ASM can describe the diverse and complex haze degree conditions in the real world more precisely. And the form of FNH-ASM is shown in (6):

$$I(x,y) = J(x,y)e^{-\beta(x,y)d(x,y)} + A(x,y)\left(1 - e^{-\beta(x,y)d(x,y)}\right) \quad (6)$$

where $I(x,y)$ is the hazy image. $J(x,y)$ is the haze-free image. $\beta(x,y)$ is an angular scattering coefficient (ASC). $d(x,y)$ is depth, which denotes the distance from the object to the camera. $A(x,y)$ is atmospheric light factor (ALF). After estimating $\beta$, $d$ and $A$, it can calculate clean image $J$ by (7):

$$J(x,y) = (I(x,y) - A(x,y))e^{\beta(x,y)d(x,y)} + A(x,y) \quad (7)$$

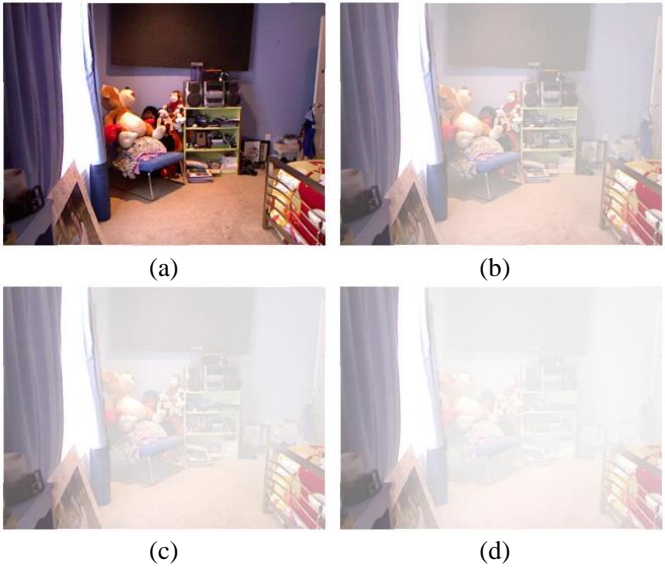

(a)           (b)

(c)           (d)

**Fig. 3.** A set of hazy images synthesized from one same haze-free image with different angular scattering coefficients. **(a)** Haze-free image. **(b)-(d)** Synthetic hazy images with angular scattering coefficient (ASC) $\beta = 0.35, 0.55$ and $0.75$, respectively.

In FNH-ASM, the ASC $\beta$ represents the concentration of haze in a hazy image. In Fig. 3, it shows a set of hazy images which are synthesized from one same haze-free image but have different angular scattering coefficients. In this set of images, the larger ASC in the image, the more severe the scene is blocked by the haze, which means the haze concentration of the image is higher. On the contrary, with the decrease of ASC, the haze will become thinner and the haze shielding phenomenon will be weakened.

Haze residue and artifacts reflect the distortion of dehazing results. To analyze the correlation between the bias of dehazing output and the bias of FNH-ASM parameters, we denote that:

$$A_{est} = A_{gt} + \Delta A \quad (8)$$

$$\beta_{est} = \beta_{gt} + \Delta\beta \quad (9)$$

$$d_{est} = d_{gt} + \Delta d \quad (10)$$

$$J_{est} = J_{gt} + \Delta J \quad (11)$$

where $A_{gt}$, $\beta_{gt}$, $d_{gt}$, $J_{gt}$ denote true ALF, ASC, depth and haze-free image, respectively. $A_{est}$, $\beta_{est}$, $d_{est}$, $J_{est}$ are the corresponding estimated values. $\Delta A$, $\Delta t$, $\Delta d$, $\Delta J$ are the bias of each parameter. From (7), (8), (9), (10) and (11), we can obtain:

$$
\begin{aligned}
\Delta J &= J_{est} - J_{gt} = (I - A_{est})e^{\beta_{est}d_{est}} + A_{gt} + \Delta A \\
&\quad - (I - A_{gt})e^{\beta_{gt}d_{gt}} - A_{gt} \\
&= (I - A_{gt} - \Delta A)e^{(\beta_{gt} + \Delta\beta)(d_{gt} + \Delta d)} \\
&\quad - (I - A_{gt})e^{\beta_{gt}d_{gt}} + \Delta A
\end{aligned} \quad (12)
$$

If $\Delta\beta = 0$ and $\Delta d = 0$, from (12), we have:

$$\Delta J = \left(1 - e^{\beta_{gt}d_{gt}}\right)\Delta A \quad (13)$$

If $\Delta A = 0$ and $\Delta d = 0$, from (12), we have:

$$\Delta J = (I - A_{gt})\left(e^{d_{gt}\cdot\Delta\beta} - 1\right)e^{d_{gt}\cdot\beta_{gt}} \quad (14)$$

As $d$ and $\beta$ have the same form and position in FNH-ASM, so if $\Delta A = 0$ and $\Delta\beta = 0$, from (12), it has same relation form for $\Delta J$ and $\Delta d$ as the relationship of $\Delta J$ and $\Delta\beta$ (equation (14)):

$$\Delta J = (I - A_{gt})\left(e^{\beta_{gt}\cdot\Delta d} - 1\right)e^{\beta_{gt}\cdot d_{gt}} \quad (15)$$

To better understand (13), we assign $d_{gt} = 1.0$, $\beta_{gt} = 0.35$, $I = 0.8$ to plot the correlation figure of $A_{gt}$, $\Delta A$ and $\Delta J$ (see Fig. 4 (a)). Similarly, to better understand (14), we assign $A_{gt} = 1.0$, $d_{gt} = 1.0$, $I = 0.8$, to plot the correlation figure of $\beta_{gt}$, $\Delta\beta$ and $\Delta J$ (see Fig. 4 (b)). As the relationship of d and $J$ is similar to $\beta$ and $J$, the correlation figure of $d_{gt}$, $\Delta d$ and $\Delta J$ can also present as Fig. 4 (b).

As for the effect of the bias estimated parameters on the dehazing result, from (13), (14), (15), Fig. 4 (a) and 4 (b), we can find out that:

1) It is a linear relationship between the bias of A (ALF) and the bias of dehazing result, but the effect of the bias of $\beta$ (ASC) or $d$ (depth) on the bias of dehazing result is nonlinear.

2) The positive and negative of ALF estimation bias has not been seen effects on the dehazing results, but it is different for ASC and depth. Compared with the negative bias of ASC and depth, positive bias causes severer biases on dehazing outputs.

3) When the estimation bias is the same, the value of ALF itself has not been seen effects on the bias of dehazing result. However, the value of ASC and depth themselves will affect the deviation of dehazing results. When the value of ASC or depth is very large(in the heavy hazy scene), the biases of ASC

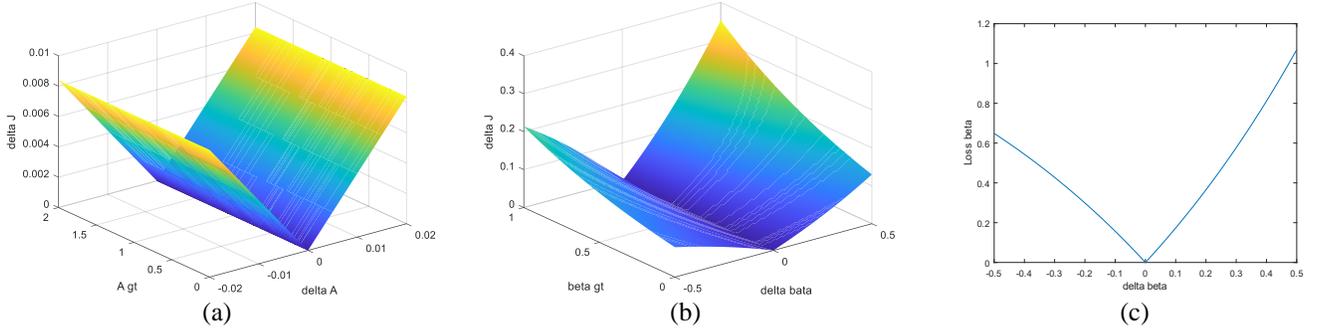

**Fig. 4.** **(a)** The correlation of the true atmospheric light factor (ALF) $A_{gt}$, the bias of estimated ALF $\Delta A$ and the bias of dehazing output $\Delta J$. It is a linear relationship between the bias of ALF and the bias of dehazing result. **(b)** The correlation of the true angular scattering coefficient (ASC) $\beta_{gt}$, the bias of estimated ASC $\Delta\beta$ and the bias of dehazing output $\Delta J$. It is a nonlinear relationship between the bias of ASC and the bias of dehazing result. The positive bias of ASC and the bias when ASC itself larger have greater impacts on dehazing results. **(c)** The plot of our proposed loss function, $\beta$-Loss. The loss reaches down to the minimum when the estimated angular scattering coefficient (ASC) value $\beta_{est}$ equals to the value of ground truth ASC $\beta_{gt}$.

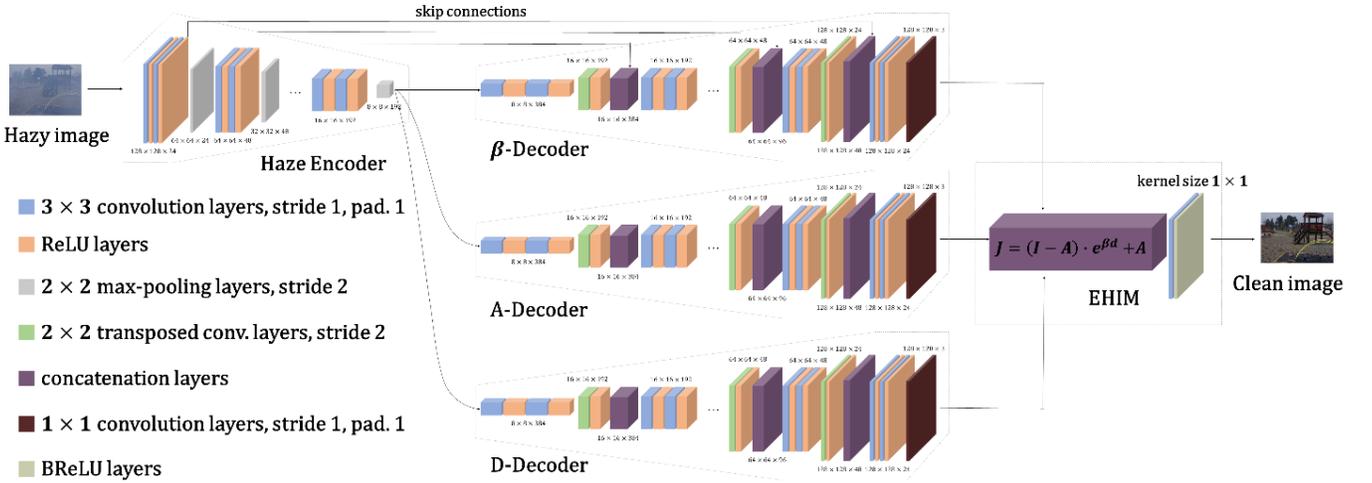

**Fig. 5.** The architecture of FNHD-Net.

or depth have a greater impact on dehazing result and cause artifacts more easily.

Therefore, our strategy for restraining dehazing bias as:

1) It can reduce the bias of dehazing results by imposing heavier penalties on the positive bias and the bias of large ASC and depth values. So, we will design two cost sensitive loss functions in section III. B to be the optimization objective of ASC and depth estimating module, respectively.

2) It is suitable for imposing a linear uniform penalty on different values of ALF, so instead of cost sensitive loss functions, the loss functions with uniform penalties for biases are appropriate to be the optimization objective of ALF estimating modules, such as MSE loss function or FWB-Loss.

### B. β-Loss and D-Loss

While using L1 or MSE loss function to learn ASC maps, it only considers the bias of ASC. But ASC is only an intermediate variable of dehazing, the ultimate goal is to make the dehazing image $J_{est}$ consistent with the ground truth haze-free image $J_{gt}$. So, we mainly consider the bias of dehazing output in our ASC training loss function. As the symbol of ASC is $\beta$ in FNH-ASM, the loss function of training the estimating ASC module is termed as β-Loss. Based on the above motivations, we set $Loss_\beta = \Delta J$. From (7) and (11), we can get:

$$
\begin{aligned}
Loss_\beta &= \Delta J = J_{est} - J_{gt} \\
&= (I - A) \cdot ((e^d)^{\beta_{est}} - (e^d)^{\beta_{gt}})
\end{aligned}
$$
(16)

$(I - A)$ is irrelevant to ASC $\beta$ as well as always positive, which means it will not change the optimum point position of the ASC estimator. So, we omit it and get a simplified loss function:

$$
Loss_\beta(\beta_{est}, \beta_{gt}) = \frac{1}{N}\sum_{x=1}^{N}|\lambda_1{}^{\beta_{est}} - \lambda_1{}^{\beta_{gt}}|
$$
(17)

where $x$ is pixel index, $N$ represents image size, the absolute value is taken to ensure $\beta$-Loss has a minimum value. $\lambda_1 = e^d$, and $d$ will be set to the mean depth of the dataset in actual implementation. For better understanding, we plot this loss function in Fig. 4 (c). We can find out from (17) and Fig. 4 (c) that $\beta$-Loss reaches down to the minimum when the value of $\beta_{est}$ is equal to the value of $\beta_{gt}$. At the same time, to restrain positive bias and the bias of large ASC, $\beta$-Loss can give a greater penalty to the positive bias than the negative one, and impose heavier punishment to the bias of large ASC than the small one.

As the relationship of depth $d$ and dehazing result $J$ is similar to ASC $\beta$ and $J$, imitating $\beta$-Loss (equation (17)), the loss function for training depth estimating module (D-Loss) can be represented as:

$$Loss_d(d_{est}, d_{gt}) = \frac{1}{N} \sum_{x=1}^{N} \left| \lambda_2^{d_{est}} - \lambda_2^{d_{gt}} \right| \qquad (18)$$

where $x$ is pixel index, $N$ represents image size, the absolute value is taken to ensure D-Loss has a minimum value. $\lambda_2 = e^{\beta}$, and $\beta$ will be set to the mean ASC of the dataset in actual implementation. As the forms of D-Loss and $\beta$-Loss are identical, their properties are the same as $\beta$-Loss. Specifically, D-Loss reaches down to the minimum when the value of $d_{est}$ is equal to the value of $d_{gt}$. And it can give a greater penalty to the positive bias than the negative one, and impose heavier punishment to the bias of large depth than the small one. Such properties ensure D-Loss having the ability for restraining the positive bias of depth and the bias of large depth. And it is helpful to reduce the bias of final dehazing results caused by parameter estimation bias.

### C. Fully non-homogeneous dehazing network (FNHD-Net)

According to FNH-ASM (equation (6)), there is a pixel-to-pixel mapping relationship between hazy image $I$ and ASC $\beta$. And it is suitable for modeling this mapping by an encoder-decoder model. Specifically, we use a U-Net [8, 9] architecture with multi-scale skip connections between the encoder and decoder. For hazy image $I$ and depth $d$, or $I$ and ALF $A$, there are similar pixel-to-pixel mapping relationships between those two variables, and is also suitable for modeling the mappings by U-Net. Based on the above motivations, we design an end-to-end dehazing network based on FNH-ASM, termed as fully non-homogeneous dehazing network (FNHD-Net). For restraining the haze residue and artifacts of dehazing result, we apply $\beta$-Loss and D-Loss to our dehazing network. The architecture of FNHD-Net is shown in Fig. 5. This network has one common encoder for converting the input haze image into a fixed length vector, and uses three decoders to transform the vector into ALF, ASC and depth map, respectively. Then, it inputs the original haze image, ALF, ASC and depth maps into estimating haze-free image module (EHIM), after adjusting and combining, finally output dehazing results. Hence, FNHD-Net has 3 parts to combine: 1) one common encoder (Haze Encoder); 2) three parallel decoders (A-Decoder, D-Decoder and $\beta$-Decoder); 3) one estimating haze-free image module (EHIM). The detail of FNHD-Net is described in below.

**Haze Encoder:** It is a 4-level encoder unit that is responsible for extracting a multi-scale latent representation of input hazy image. At the first level of Haze Encoder, the convolution layers have 24 channels. For each subsequent level, the number of channels is doubled (i.e., the fourth level has 192 channels for each convolution layer).

**A-Decoder, D-Decoder and $\beta$-Decoder:** They are three parallel 4-level decoders with same structure. More specifically, they all have a bottleneck and transposed convolutional layers to realize the deconvolution of the encoder's output. A-Decoder, D-Decoder and $\beta$-Decoder directly learn the mapping from haze image to ALF, depth and ASC, then output ALF map, depth map and ASC map, respectively. Contrary to the structure of Haze Encoder, the number of channels of decoders decrease with the deepening of network layers. At the first level of the decoders, the convolution layers have 192 channels. For each subsequent level, the number of channels is halved (i.e., the fourth level has 24 channels for each convolution layer).

**Estimate haze-free image module (EHIM):** After obtaining ALF $A$, ASC $\beta$ and depth $d$, (7) is employed to figure out the haze-free image $J$. To refine the output and remove noise automatically, we process $J$ with a convolution layer followed by a BReLU [33] activation function layer.

Our proposed FNHD-Net is delicately designed based on FNH-ASM for single image dehazing. And it can restrain haze residue and artifacts by finely estimating FNH-ASM parameters pixel by pixel.

## IV. EVALUATIONS

We conduct detailed experiments to evaluate the proposed method on three commonly used image dehazing datasets: NYU-Depth [34], O-Haze [35] and NH-Haze [36]. In this section, we first introduce the datasets and implementation details. Then, we present the evaluation of the restraining bias capability of FNHD-Net. Next, we compare the state-of-the-art methods. Finally, we conduct ablation studies to evaluate the effectiveness of the proposed FNH-ASM, $\beta$-Loss and D-Loss, respectively.

### A. Datasets

**NYU-Depth v2 [34]:** Adequate training data is essential for CNN-based methods, but since it is very difficult to collect both clean and hazy images of the same scene, real world dehazing databases are very scarce. Lots of classical dehazing methods, such as AOD-Net [15], FWB-Net [4], use ASM [1] to synthesize dehazing datasets. By convention, we use NYU-Depth [34] to synthesize our dataset based on FNH-ASM. NYU-Depth v2 provides 1449 clean images and their corresponding depth information. Referring to FNH-ASM, we synthesize 3 datasets of 3 different haze concentrations on NYU-Depth v2. The synthetic method is setting a basic ASC $\beta$ on each dataset and apply a random disturbance of less than 100% to the ASC of every pixel (i.e., when the basic ASC of the dataset is 0.35, the range of ASC of every pixel of the dataset are from 0 to 0.35). The basic ASC $\beta$ of each dataset is 0.35, 0.55 and 0.75, respectively. Among them, the haze concentration in the dataset increases with the increase of ASC. And for each dataset contains 1000 images and the rest 449 images for training and testing set, respectively.

At the same time, we randomly set the basic ALF between 0.3 and 1.5 for each image of each dataset, and apply a random

| β | ASC | | Depth | |
|---|---|---|---|---|
| | AB ($\times 10^{-3}$) | PoPB (%) | AB (m) | PoPB (%) |
| 0.35 | -4.75 | 20.3 | -0.124 | 27.4 |
| 0.55 | -6.09 | 18.6 | -0.132 | 25.5 |
| 0.75 | -8.37 | 17.2 | -0.135 | 29.8 |

**Note:** AB, PoPB denotes average bias, the proportion of positive bias, respectively. The unit of depth is meter.

The average biases of both ASC and depth of all the testing datasets are tiny values, most of them have negative biases of ASC and depth. Moreover, with the increase of haze degree, the proportion of positive bias of ASC decrease.

disturbance of less than 20% to the ALF of each pixel. Using this method, we synthesize the heterogeneous atmospheric light map of each synthetic hazy image.

**O-Haze** [35] is a real-world homogeneous hazy dataset for image dehazing. It consists of 40 training images and 5 testing images. All the hazy and ground truth haze-free images are taken with live-action. And all the hazy images contain haze with uniform concentration. As this dataset contains fewer images, it is often used to fine-tune the weight and verify the performance of the model.

**NH-Haze** [36] is a real-world heterogeneous hazy dataset for image dehazing. It consists of 47 training images and 8 testing images. All the hazy and ground truth haze-free images are taken with live-action, and it is often used to fine-tune and verify dehazing models, similarly. Compared with O-Haze, it involves hazy images of uneven haze concentration.

### B. Implementation details

In the training session, we train FNHD-Net on all the 3 NYU-Depth-based training datasets with different haze degrees, and use β-Loss, D-Loss and FWB-Loss [4] to train β-Decoder, D-Decoder and A-Decoder under the supervision of ground truth ASC, depth and ALF maps, respectively. The $\lambda_1$ in β-Loss is 5.212 on all the three training sets. And the $\lambda_2$ in D-Loss is 1.195, 1.290, 1.471 on the training set with basic $\beta$ of 0.35, 0.55 and 0.75, respectively. Then, we use MSE loss to train the whole FNHD-Net, under the supervision of ground truth haze-free images. With the help of stochastic gradient descent (SGD) algorithm [37], using a learning rate of 1e-4, a weight decay of 1e-4, and a momentum of 0.9, we take 100 iterations to train β-Decoder, D-Decoder and A-Decoder in parallel, and take another 100 iterations to train the whole network, then repeat the cycle until all the three decoders and FNHD-Net converge. While in the testing session, hazy images are input into the model and clean images are used for evaluation. It should be noted that we only test the performance of FNHD-Net on the testing set with the same haze degree as the training set.

In addition to NYU-Depth v2, we also verify the performance of FNHD-Net on two real-world datasets, O-Haze and NH-Haze. Where O-Haze is a real homogeneous hazy dataset and NH-Haze is a real heterogeneous hazy dataset. For fair comparison, before verifying the performance of our model on these two datasets, we first fine-tune the pre-trained model on the training sets of corresponding datasets. And the used pre-trained model weight is the model trained on NYU-Depth v2 with β = 0.55.

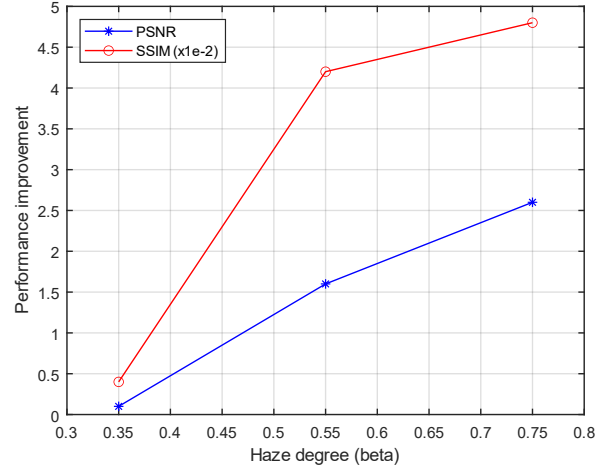

**Fig. 6.** The performance improvement of FNHD-Net to the suboptimal performance comparison method on NYU-Depth v2 with different haze degrees. With the increase of haze degree, the performance improvement of FNH-ASM increase, both PSNR and SSIM have risen.

All our experiments are conducted by PyTorch [38] on a GPU of GeForce GTX TITAN X. And the CPU configuration is Intel(R) Xeon(R) CPU E5-2680 v4 @ 2.40GHz.

### C. Evaluation of restraining bias capability of FNHD-Net

To evaluate the restraining capability of FNHD-Net to the dangerous positive bias of ASC, we collect the output ASC maps from β-Decoder of all the testing samples of NYU-Depth v2 with all the 3 concentrations of haze. Comparing with the ground truth ASC maps, we calculate the bias of each pixel on each sample pair to get an average bias. The average bias and the proportion of positive bias of all the 3 testing datasets are shown on Table I. The average biases of both ASC and depth of all the testing datasets are tiny values. At the same time, most of them have a negative bias of ASC and depth. It shows that our FNHD-Net can restrain large bias and dangerous positive bias of ASC and depth effectively with the help of β-Loss and D-Loss.

Moreover, with the increase of haze degree, the proportion of positive bias of ASC decreases, which means FNHD-Net has a stronger ability to restrain the dangerous positive bias in dense haze scenes. This characteristic of FNHD-Net is also helpful to output high-quality dehazing results.

### D. Evaluations on NYU-Depth v2 dataset

We evaluate FNHD-Net on 3 synthetic hazy datasets, which are synthesized from NYU-Depth v2 [34]. The dataset synthesis method can be seen in Section IV. A, and the basic ASC β of each dataset is 0.35, 0.55 and 0.75, respectively. Another eleven high-performance methods: DCP [10], Berman [12], DehazeNet [13], AOD-Net [15], DCPDN [21], GDN [18],

TABLE II
Quantitative comparisons for different methods

| Model | NYU-Depth v2 | | | | | | O-Haze | | NH-Haze | |
|---|---|---|---|---|---|---|---|---|---|---|
| | $\beta = 0.35$ | | 0.55 | | 0.75 | | - | | - | |
| | PSNR | SSIM | PSNR | SSIM | PSNR | SSIM | PSNR | SSIM | PSNR | SSIM |
| DCP [10] | 17.3 | 0.785 | 16.3 | 0.743 | 13.5 | 0.656 | 16.6 | 0.735 | 12.9 | 0.472 |
| Berman [12] | 16.6 | 0.763 | 13.2 | 0.694 | 10.3 | 0.613 | 16.6 | 0.750 | 12.5 | 0.530 |
| DehazeNet [13] | 14.7 | 0.749 | 10.2 | 0.630 | 7.8 | 0.546 | 16.2 | 0.666 | 16.6 | 0.524 |
| AOD-Net [15] | 12.8 | 0.688 | 10.1 | 0.602 | 8.5 | 0.539 | 17.1 | 0.664 | 15.4 | 0.569 |
| DCPDN [21] | 12.6 | 0.656 | 10.3 | 0.534 | 9.0 | 0.434 | 13.9 | 0.686 | 17.1 | 0.585 |
| GDN [18] | 18.9 | 0.763 | 13.3 | 0.617 | 10.2 | 0.521 | 18.7 | 0.672 | 13.8 | 0.537 |
| EPDN [22] | <u>20.5</u> | <u>0.808</u> | <u>16.8</u> | 0.728 | 14.4 | 0.633 | 16.6 | 0.690 | - | - |
| MSCNN-HE [14] | 13.2 | 0.705 | 9.9 | 0.583 | 7.9 | 0.471 | 19.0 | 0.675 | - | - |
| FD-GAN [23] | 18.4 | 0.752 | 16.6 | 0.692 | <u>14.6</u> | 0.600 | 18.1 | 0.730 | - | - |
| FFA-Net [19] | 12.0 | 0.629 | 10.1 | 0.512 | 9.1 | 0.440 | - | - | <u>19.9</u> | <u>0.692</u> |
| FWB-Net [4] | 18.9 | 0.794 | 15.6 | <u>0.744</u> | 13.9 | <u>0.659</u> | <u>19.7</u> | <u>0.763</u> | - | - |
| **Ours** | **20.6** | **0.812** | **18.4** | **0.786** | **17.2** | **0.707** | **20.1** | **0.776** | **22.3** | **0.743** |
| Improvement | 0.1 | 0.004 | 1.6 | 0.042 | 2.6 | 0.048 | 0.4 | 0.013 | 2.4 | 0.051 |

**Note: Boldface** and <u>Underlined</u> represent the optimal and suboptimal performance respectively. "Improvement" means the performance improvement of the optimal performance comparison method to the suboptimal.

Our method ranks both first of PSNR and SSIM on the whole 3 datasets.

EPDN [22], MSCNN-HE [14], FD-GAN [23], FFA-Net [19] and FWB-Net [4] are reimplemented on the training set of all the 3 datasets for comparison. The qualitative and quantitative results are shown in and Fig. 7 and Table II.

For qualitative comparisons, the results of image prior-based methods suffer from color shift in heavy hazy areas or white image patches, and there are obvious artifacts on it. But nine deep learning-based methods avoid color shift and have fewer artifacts. However, they produce poor-effect dehazing outputs. The most obvious is there are quantities of haze residue in heavy hazy areas. However, the results of our FNHD-Net are haze-free as well as avoiding color shift, artifacts and haze residue. It shows our method has the most outstanding performance of all the compared methods.

Although our work focuses on improving visibility, it also does well in quantitative comparisons. Peak Signal to Noise Ratio (PSNR) [39] and Structural Similarity Index Measure (SSIM) [40] are widely used in evaluating dehazing results. We calculate PSNR and SSIM of all compared methods and our method for providing the result in Table II. Our FNHD-Net ranks first in both indexes on all the 3 synthetic datasets. It implies that our proposed method is effective.

Moreover, with the increase of haze concentration, our method and other comparison dehazing methods have certain performance degradation, but the performance improvement of our method is improved compared with the other methods. Specifically, Table II and Fig. 6 shows that the performance improvement of FNHD-Net is 0.1, 1.6, 2.6 (for PSNR) and 0.004, 0.042, 0.048 (for SSIM) compared with the suboptimal dehazing method on the dataset with $\beta$ of 0.35, 0.55 and 0.75, respectively. It shows the performance improvement of FNHD-Net increases more obviously with the increase of haze concentration compared with other dehazing methods, which means that our method has greater advantages in removing dense haze.

### E. Evaluations on O-Haze dataset

We evaluate our method and eleven comparing methods: DCP [10], Berman [12], DehazeNet [13], AOD-Net [15], DCPDN [21], GDN [18], EPDN [22], MSCNN-HE [14], FD-GAN [23], FFA-Net [19] and FWB-Net [4] on real-world homogeneous hazy image dataset O-Haze [35] and present some of the qualitative and quantitative results in Fig. 8 and Table II, respectively. For fair comparison, the quantitative results of {FWB-Net [4], DCP [10], Berman [12], DehazeNet [13], AOD-Net [15], GDN [18]}, {DCPDN [21], EPDN [22], GAN [23]} and [14] on O-Haze come from [4], [23] and [35], respectively. And all the comparison methods are reimplemented on the O-Haze training set for getting qualitative results.

In Fig. 8, it can be seen that our results not only are the closest to the ground truth, but also performs best in avoiding color shift and reducing haze residue. After dehazing, the results of the compared methods are blue, while ours restore the color. Moreover, in Table II, our method performs the best in PSNR and SSIM on the O-Haze dataset. Overall, our proposed method restores more details and obtains visually pleasing images in real world homogeneous hazy conditions.

### F. Evaluations on NH-Haze dataset

We evaluate our method and eleven comparing methods DCP [10], Berman [12], DehazeNet [13], AOD-Net [15], DCPDN [21], GDN [18], EPDN [22], MSCNN-HE [14], FD-GAN [23], FFA-Net [19] and FWB-Net [4] on real-world heterogeneous hazy image dataset NH-Haze [36] and present some of the qualitative and quantitative results in Fig. 9 and Table II, respectively. For fair comparison, the quantitative results of {DCP [10], Berman [12], DCPDN [21]} and {DehazeNet [13], AOD-Net [15], GDN [18], FFA-Net [19]} on NH-Haze come from [36] and [41], respectively. And all the comparison methods are reimplemented on the NH-Haze training set for getting qualitative results. As this paper mainly solves the problem of haze residue effect on dehazing results in heterogeneous haze scenes, so the performance comparison results on this dataset are the most convincing.

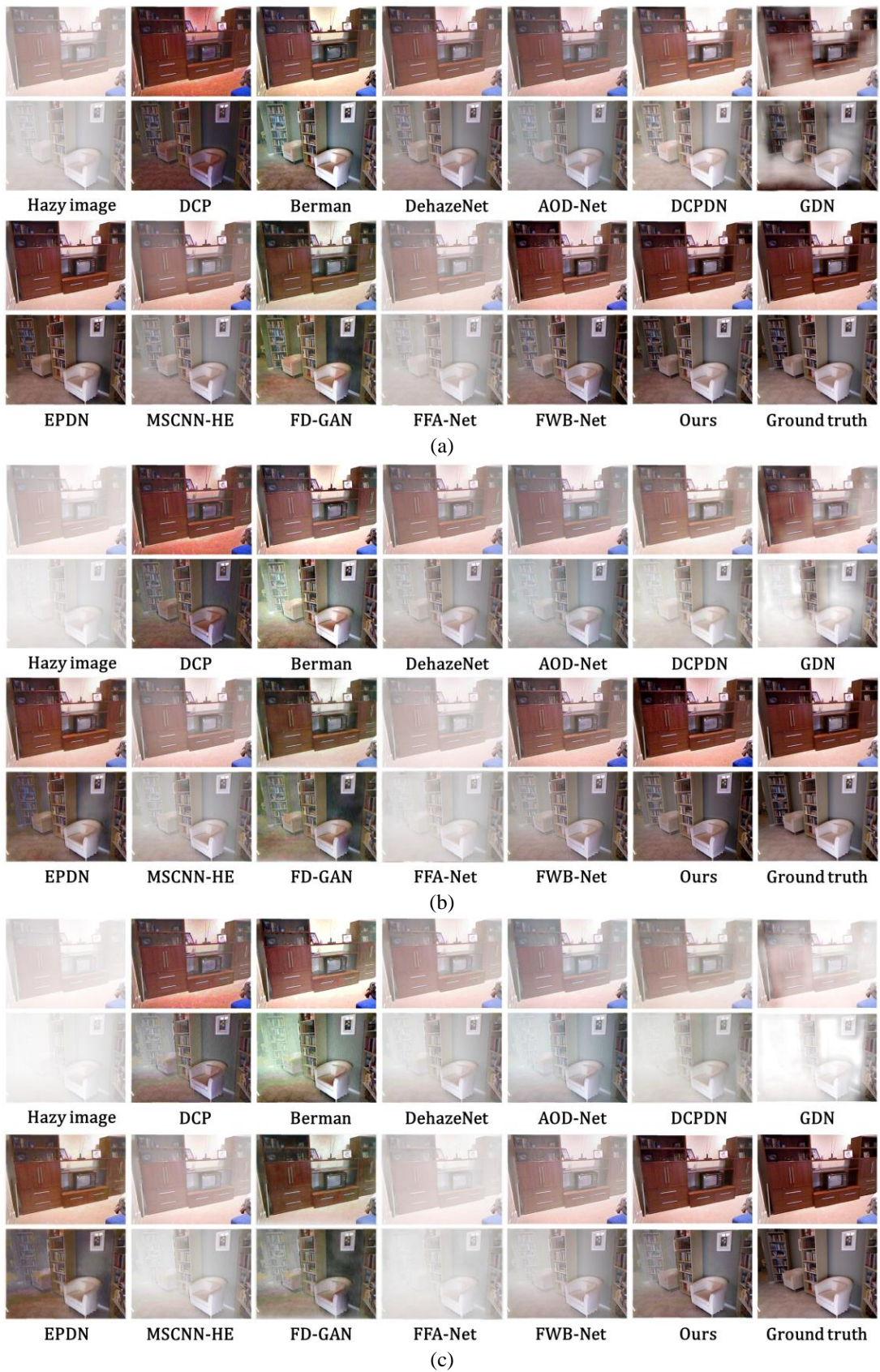

**Fig. 7. (a)-(c)** Evaluation on synthetic hazy dataset NYU-Depth with $\beta = 0.35$, 0.55 and 0.75, respectively. Results demonstrate that our method can get clean dehazing results with restraining haze residue, artifacts and color shift. While others get unclean dehazing results, appear artifacts or suffer from color shift. And the performance improvement of our method increases more obviously with the increase of haze concentration compared with other dehazing methods.

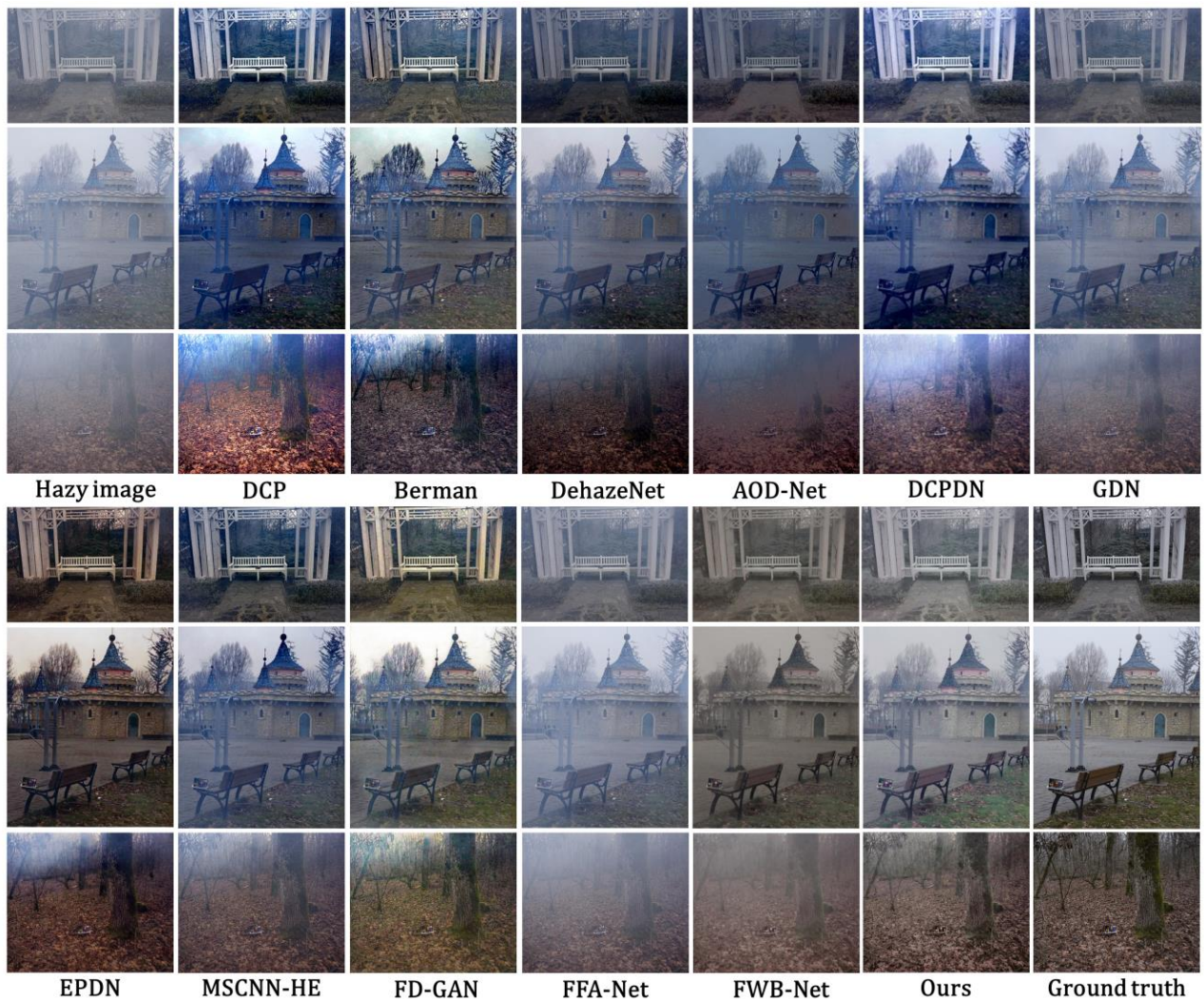

**Fig. 8.** Evaluation on synthetic hazy dataset O-Haze. Results demonstrate that our method can get clean dehazing results with restraining haze residue, artifacts and color shift. While others get unclean dehazing results, appear artifacts or suffer from color shift.

In Fig. 9, it can see that our results also avoid haze residue and color shift while removing haze. After dehazing, the results of the compared methods turn to blue or yellow as well as with extensive unremoved haze, while ours restore the color and almost completely remove haze. Simultaneously, in Table II, our method performs the best in PSNR and SSIM on the NH-Haze dataset. Therefore, the qualitative and quantitative results show that our method leads to superior performance in real world heterogeneous hazy conditions.

Moreover, in Table II, the performance improvement of our method is higher on NH-Haze than on O-Haze. As NH-Haze is a heterogeneous haze dataset while O-Haze only contains homogeneous haze, such phenomenon indicates that our FNHD-Net has more superiority in heterogeneous haze scene.

### G. Ablation Studies

We conduct several ablation experiments on NYU-Depth v2 with all the 3 concentrations of haze to analyze the proposed FNH-ASM, β-Loss and D-Loss. Details are discussed as follows.

**Effectiveness of FNH-ASM.** We devise a comparative experiment to demonstrate the effectiveness of FNH-ASM. We replace the β-Decoder and D-Decoder of FNHD-Net with the estimating transmission module used in ASM and NH-ASM based dehazing models, and retrain the model. For fair comparison, the replaced estimating transmission module is also set as a U-Net of the same layer number, size and structure as Haze Encoder and β-Decoder.

The experiment results in Table III (a) and (b) and Fig. 10 show that FNH-ASM brings about an obvious ascent in performance. The PSNR of module based on FNH-ASM improves by 0.7, 1.0 and 1.2 compared with the dehazing model based on NH-ASM or ASM on the testing datasets with $\beta$ of 0.35, 0.55 and 0.75, respectively. It also shows that the

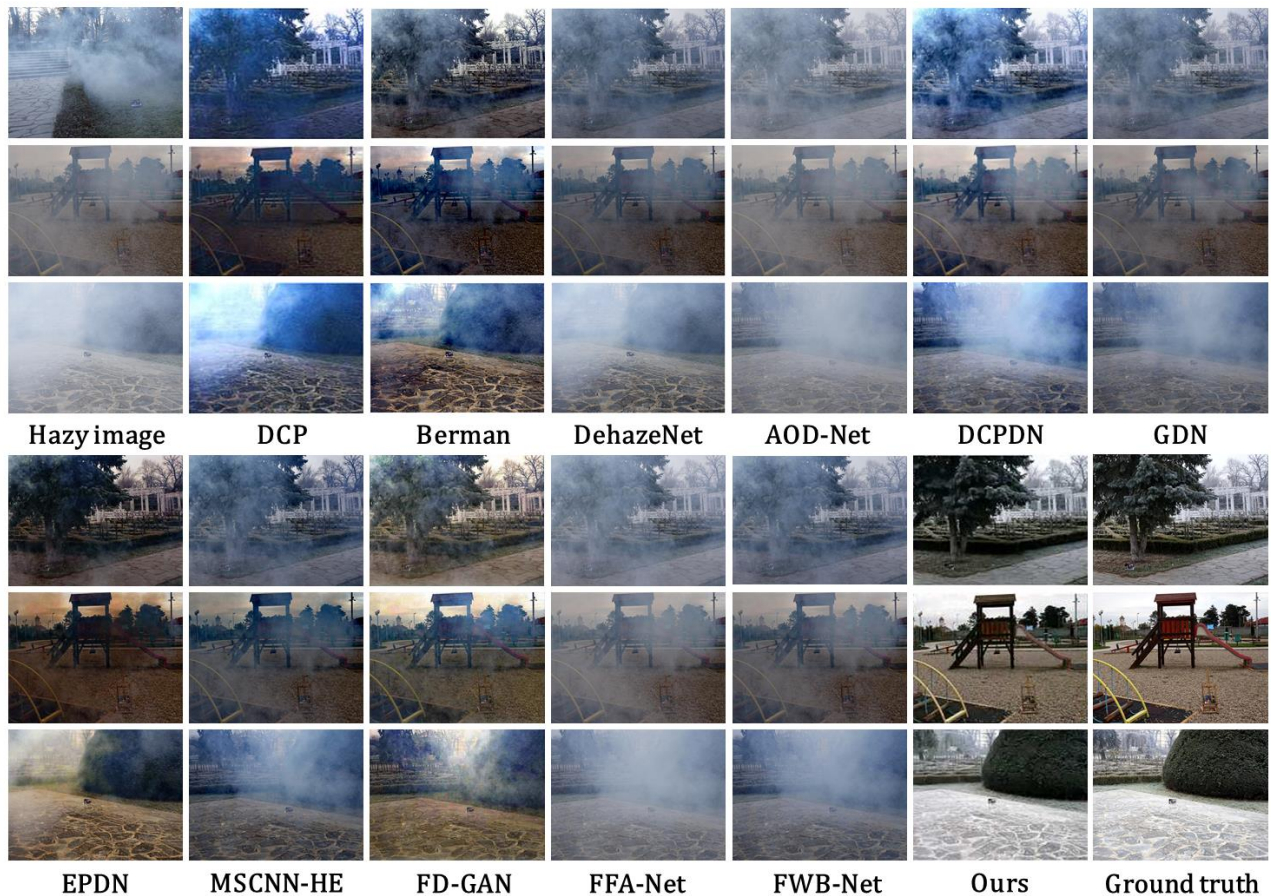

| Hazy image | DCP | Berman | DehazeNet | AOD-Net | DCPDN | GDN |

| EPDN | MSCNN-HE | FD-GAN | FFA-Net | FWB-Net | Ours | Ground truth |

**Fig. 9.** Evaluation on synthetic hazy dataset NH-Haze. Results demonstrate that our method can get clean dehazing results with restraining haze residue, artifacts and color shift. While others get unclean dehazing results, appear artifacts or suffer from color shift.

TABLE III
ABLATION STUDIES ON NYU-DEPTH V2

| Experiment | | (a) | (b) | (c) | (d) | (e) |
|---|---|---|---|---|---|---|
| FNH-ASM | | | √ | √ | √ | √ |
| β-Loss | | | | √ | | √ |
| D-Loss | | | | | √ | √ |
| β | 0.35 | P | 18.9 | 19.6 | 20.3 | 20.0 | 20.6 |
| | | G | - | 0.7 | 0.7 | 0.4 | 1.0 |
| | 0.55 | P | 15.6 | 16.6 | 17.7 | 17.2 | 18.4 |
| | | G | - | 1.0 | 1.1 | 0.6 | 1.8 |
| | 0.75 | P | 13.9 | 15.1 | 16.5 | 15.6 | 17.2 |
| | | G | - | 1.2 | 1.4 | 0.5 | 2.1 |

**Note:** P, G, denotes PSNR, the performance gain, respectively.

The performance gain in experiment (b) is compared with that in experiment (a), and the performance gain in experiment (c) - (e) are compared with that in experiment (b).

performance gain of FNH-ASM to ASM increases with the increase of ASC, which means our proposed FNH-ASM has a stronger performance advantage in dense haze scenes.

**Effectiveness of β-Loss and D-Loss.** We devise a set of comparative experiments to demonstrate the effectiveness of β-Loss and D-Loss. We set four training strategies to train FNHD-

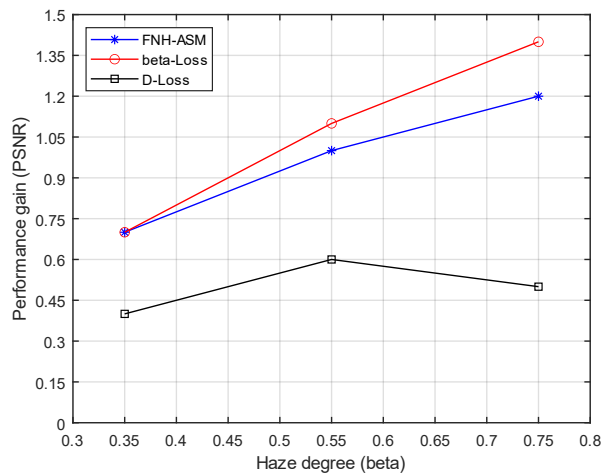

**Fig. 10.** The performance gain of FNH-ASM, β-Loss and D-loss on different haze degree scenes. With the increase of haze degree, the performance gain of FNH-ASM and β-Loss increase, while the performance gain of D-Loss is almost the same.

Net: using MSE loss to replace both β-Loss and D-Loss, using MSE loss to replace D-Loss only and keeping β- Loss, using MSE loss to replace β-Loss only and keeping D-Loss, keeping both β-Loss and D-Loss.

The experiment results in Table III (b) - (e) and Fig. 10 show that both β-Loss and D-Loss bring ascents in performance. The PSNR of the module trained by β-Loss improves by 0.7, 1.1 and 1.4 compared with the model trained by MSE loss on the datasets with $\beta$ of 0.35, 0.55 and 0.75, respectively. At the same time, The PSNR of the module trained by D-Loss improves by 0.4, 0.6 and 0.5 compared to the model trained by MSE loss on the datasets with $\beta$ of 0.35, 0.55 and 0.75, respectively. It shows that:

1) With the increase of ASC, the performance gain of β-Loss increases, which means that our proposed β-Loss has a stronger performance advantage in dense haze scenes.

2) With different ASCs, the performance gain of D-Loss is almost the same, which means that D-Loss has the same effectiveness for different concentrations of haze.

3) With the same ASC, the performance gain of β-Loss is always more than it of D-Loss. It means that β-Loss is more effective than D-Loss in improving the quality of dehazing results, which also proves that accurate estimation of ASC is the key to eliminate haze residual effect.

## V. CONCLUSIONS

In this paper, first, we propose a new fully non-homogeneous atmospheric scattering model (FNH-ASM) to precisely model the complex non-homogeneous haze degree conditions in real hazy scenes. Then, we analyze the influence of parameter estimation bias in FNH-ASM on the dehazing result bias. It shows that the positive bias of the estimated angular scattering coefficient (ASC) and the bias of large ASC value will cause serious distortion on the dehazing result. It also has the same conclusion for depth. Based on these discoveries, we propose a novel β-Loss and a novel D-Loss for estimating ASC map and depth map, respectively. We further incorporate the proposed β-Loss and D-Loss to form a unified dehazing network called fully non-homogeneous dehazing network (FNHD-Net), which is designed based on FNH-ASM with the ability of restraining haze residue and getting clear dehazing outputs. Experimental results show our method can get outstanding clean dehazing outputs without haze residue. And the performance improvement of our method increases more obviously with the increase of haze concentration compared with other dehazing methods. Moreover, in heterogeneous and dense haze scenes, our proposed method also has more superiority.

## REFERENCES

[1] E. J. McCartney, *Optics of the atmosphere: scattering by molecules and particles*, John Wiley and Sons, New York, 1976.

[2] S. G. Narasimhan, and S. K. Nayar, "Vision and the atmosphere," *International Journal of Computer Vision,* Springer, pp. 48(3): 233–254, 2002.

[3] Y. Li, S. You, M. S. Brown, and R. T. Tan, "Haze visibility enhancement: A survey and quantitative benchmarking," *Computer Vision and Image Understanding,* Springer, pp. 1(65): 1–16, 2017.

[4] C. Wang, Y. Huang, Y. Zou, and Y. Xu, "FWB-Net: Front White Balance Network for Color Shift Correction in Single Image Dehazing via Atmospheric Light Estimation," *IEEE International Conference on Acoustics, Speech and Signal Processing*, pp. 2040-2044, 2021.

[5] Y. LeCun et al, "Backpropagation Applied to Handwritten Zip Code Recognition," *Neural Computation*, pp. 1(4): 541-551, 1989.

[6] Ismail Gultepe, *Fog and Boundary Layer Clouds: Fog Visibility and Forecasting*, page 1126, Springer, New York, 2008.

[7] E. Lehmann, C. George, *Theory of Point Estimation (2nd ed.)*, Springer, New York, 1998.

[8] O. Ronneberger, P. Fischer, and T. Brox, "U-Net: Convolutional Networks for Biomedical Image Segmentation," *Medical Image Computing and Computer-Assisted Intervention*, pp. 9351: 234--241, 2015.

[9] L. Liu et al, "A survey on U-shaped networks in medical image segmentations," *Neurocomputing,* pp. 409: 244-258, 2020.

[10] K. He, J. Sun and X. Tang. "Single image haze removal using dark channel prior," *IEEE Conference on Computer Vision and Pattern Recognition*, pp. 1956-1963, 2009.

[11] Q. Zhu, J. Mai and L. Shao, "A Fast Single Image Haze Removal Algorithm Using Color Attenuation Prior," *IEEE Transactions on Image Processing,* pp. 24(11): 3522-3533, 2015.

[12] D. Berman, T. Treibitz, and S. Avidan, "Non-local Image Dehazing," *IEEE Conference on Computer Vision and Pattern Recognition*, pp. 1674-1682, 2016.

[13] B. Cai, X. Xu, K. Jia, C. Qing, and D. Tao, "DehazeNet: An End-to-End System for Single Image Haze Removal," *IEEE Transactions on Image Processing*, pp. 25(11): 5187–5198, 2016.

[14] W. Ren et al, "Single Image Dehazing via Multi-Scale Convolutional Neural Networks with Holistic Edges," International Journal of Computer Vision, Springer, pp. 128: 240–259, 2020.

[15] B. Li, X. Peng, Z. Wang, J. Xu, and D. Feng, "AOD-Net: All-in-One Dehazing Network," *IEEE International Conference on Computer Vision*, pp. 4780-4788, 2017.

[16] Z. Chen, Y. Wang and Y. Zou, "Inverse Atmospheric Scattering Modeling with Convolutional Neural Networks for Single Image Dehazing," *IEEE International Conference on Acoustics, Speech and Signal Processing,* pp. 2626-2630, 2018.

[17] W. Ren et al, "Gated Fusion network for Single Image Dehazing," *IEEE Conference on Computer Vision and Pattern Recognition*, pp. 3253-3261, 2018.

[18] X. Liu, Y. Ma, Z. Shi, and J. Chen, "GridDehazeNet: Attention-Based Multi-Scale Network for Image Dehazing," *International Conference on Computer Vision*, pp. 7313-7322, 2019.

[19] X. Qin et al, "FFA-Net: Feature Fusion Attention Network for Single Image Dehazing," *AAAI Conference on Artificial Intelligence,* pp. 11908-11915, 2020.

[20] H. Dong et al, "Multi-Scale Boosted Dehazing Network with Dense Feature Fusion," *IEEE Conference on Computer Vision and Pattern Recognition,* pp. 2154-2164, 2020.

[21] H. Zhang, V. M. Patel, "Densely Connected Pyramid Dehazing Network," *IEEE Conference on Computer Vision and Pattern Recognition,* pp. 3194-3203, 2018.

[22] Y. Qu, Y. Chen, J. Huang and Y. Xie, "Enhanced Pix2pix Dehazing Network," *IEEE Conference on Computer Vision and Pattern Recognition,* pp. 8160-8168, 2019.

[23] Y. Dong et al, "FD-GAN: Generative Adversarial Networks with Fusion-Discriminator for Single Image Dehazing," *AAAI Conference on Artificial Intelligence,* pp. 10729-10736, 2020.

[24] I. Goodfellow et al, "Generative Adversarial Networks," *International Conference on Neural Information Processing Systems,* pp. 2672–2680, 2014.

[25] I. Goodfellow, Y. Bengio, A. Courville, *Deep Learning,* The MIT Press, Cambridge MA, 2016.

[26] J. Sun et al, "SAUNet: Shape Attentive U-Net for Interpretable Medical Image Segmentation," *Medical Image Computing and Computer-Assisted Intervention,* pp. 797-806, 2020.

[27] T. Tarasiewicz et al, "Skinny: A Lightweight U-Net for Skin Detection and Segmentation," *IEEE International Conference on Image Processing,* pp. 2501-2505, 2020.

[28] W. Yu et al, "Liver Vessels Segmentation Based on 3d Residual U-NET," *IEEE International Conference on Image Processing,* pp. 250-254, 2019.

[29] L. Zhu et al, "Stacked U-shape networks with channel-wise attention for image super-resolution," *Neurocomputing,* pp. 345: 58-66, 2019.

[30] S. Yin et al, "Attentive U-recurrent encoder-decoder network for image dehazing," *Neurocomputing,* pp. 437: 143-156, 2021

[31] Y. Lee et al, "Image Dehazing With Contextualized Attentive U-NET," *IEEE International Conference on Image Processing,* pp. 1068-1072, 2020.

[32] H. Yang et al, "Wavelet U-Net and the Chromatic Adaptation Transform for Single Image Dehazing," *IEEE International Conference on Image Processing,* pp. 2736-2740, 2019.

[33] L. Eidnes, A. Nøkland, "Shifting Mean Activation Towards Zero with Bipolar Activation Functions," *arXiv preprint,* arXiv:1709.04054, 2017.

[34] N. Silberman, D. Hoiem, P. Kohli, and R. Fergus, "Indoor Segmentation and Support Inference from RGBD Images," *European Conference on Computer Vision,* pp. 746-760, 2012.

[35] C. O. Ancuti, C. Ancuti, R. Timofte, and C. D. Vleeschouwer, "O-HAZE: a dehazing benchmark with real hazy and haze-free outdoor images," *NTIRE Workshop, NTIRE CVPR'18,* 2018.

[36] C. O. Ancuti, C. Ancuti and R. Timofte, "NH-HAZE: An Image Dehazing Benchmark with Non-Homogeneous Hazy and Haze-Free Images," *NTIRE Workshop, NTIRE CVPR'20,* 2020.

[37] B. Léon, and B. Olivier, *Optimization for Machine Learning,* MIT Press, Cambridge, USA, pp. 351–368, 2012.

[38] S. Paszke, S. Gross, F. Massa, "PyTorch: An Imperative Style, High-Performance Deep Learning Library," *Neural Information Processing Systems,* pp. 8024-8035, 2019.

[39] A. Hore and D. Ziou, "Image Quality Metrics: PSNR vs. SSIM," *International Conference on Pattern Recognition,* pp. 2366-2369, 2010.

[40] Z. Wang, A. C. Bovik, H. R. Sheikh and E. P. Simoncelli, "Image quality assessment: From error visibility to structural similarity," *IEEE Transactions on Image Processing,* pp. 13(4): 600-612, 2004.

[41] H. Wu et al, "Contrastive Learning for Compact Single Image Dehazing," *IEEE Conference on Computer Vision and Pattern Recognition,* pp. 10551-10560, 2021.